\documentclass[11pt]{article}

\usepackage[utf8]{inputenc}
\usepackage[T1]{fontenc}
\usepackage{lmodern}
\usepackage{microtype}

\usepackage{amsmath,amssymb,amsfonts,amsthm}

\usepackage{graphicx}
\usepackage{booktabs}
\usepackage{multirow}
\usepackage{caption}
\usepackage{subcaption}
\usepackage{float}

\usepackage[margin=2.5cm]{geometry}
\usepackage{titlesec}
\usepackage{enumitem}
\setlist{nosep,leftmargin=1.2em}

\usepackage[ruled,vlined,linesnumbered]{algorithm2e}

\SetAlgoNoEnd
\SetKwInput{KwInput}{Input}
\SetKwInput{KwOutput}{Output}
\SetKwInput{KwHyper}{Hyperparams}

\usepackage[numbers,sort&compress]{natbib}
\usepackage[colorlinks,citecolor=blue!60!black,linkcolor=blue!60!black,urlcolor=blue!50!black]{hyperref}
\usepackage{xcolor}

\newtheorem{definition}{Definition}

\newcommand{\bphi}{\boldsymbol{\varphi}}
\newcommand{\bh}{\mathbf{h}}
\newcommand{\bz}{\mathbf{z}}
\newcommand{\R}{\mathbb{R}}
\newcommand{\kunsafe}{\kappa_{\mathrm{unsafe}}}

\newcommand{\cone}{\textsc{Cone}}
\newcommand{\poly}{\textsc{Polytope}}
\newcommand{\cyl}{\textsc{Cylinder}}

\title{\Large\bfseries Geometry-Guided Constraint Learning\\
for LLM Safety Classification}

\author{Fumiaki Uehara, Koo Imai, Masato Tsutsumi, Keigo Kansa, Sora Usui, Yuki Kobiyama}
\date{}

\begin{document}
\maketitle
\thispagestyle{empty}

\begin{center}
\begin{minipage}{0.80\textwidth}
\centerline{\textbf{Abstract}}
\vspace{0.3em}
\small
Safety as Polytope (SaP) learns linear half-space constraints in LLM
hidden space but requires per-category tuning of the constraint count
$K$.
We show that sparse autoencoder (SAE) feature extraction resolves this:
$K{=}2$ becomes optimal for 12/14 categories on Qwen3.5-9B,
achieving 96--99\% accuracy per category on our BeaverTails
classification benchmark, largely eliminating the need for exhaustive
sweeps ($K{=}4$--$25$ with random initialization).
This convergence to two planes is consistent with the Linear
Representation Hypothesis, providing suggestive evidence that safety
boundaries in this setting admit a low-dimensional linear description
in the SAE feature space.
Building on this geometric perspective, we introduce a cone constraint
whose learnable aperture adapts to each category's cluster
concentration, stabilized by a three-phase training schedule.
SAE initialization selectively benefits the cone (12/13 wins,
$p{<}0.001$) but not the polytope ($p{=}0.11$), suggesting that the
SAE-initialized cone captures the angular spread of unsafe clusters
more accurately than the flat polytope.
\end{minipage}
\end{center}
\vspace{1em}

\section{Introduction}\label{sec:intro}

Safety alignment of large language models (LLMs) remains a central
challenge.
Current approaches predominantly rely on reinforcement learning from
human feedback (RLHF)~\cite{ouyang2022instructgpt} or Constitutional
AI~\cite{bai2022constitutional}, which modify model weights globally and
offer limited interpretability regarding what the model has learned about
safety.
Moreover, these defenses can be circumvented by adversarial
prompts~\cite{zou2023gcg,wei2024jailbroken,chao2023pair}, motivating
post-hoc safety mechanisms that complement alignment training.

Recent work on the Linear Representation Hypothesis
(LRH)~\cite{park2024lrh,elhage2022superposition,nanda2023othello} has
revealed that high-level concepts---including safety-relevant ones---are
encoded as geometric structures in LLM representation spaces.
Building on this insight, \citet{chen2025sap} introduced the Safety
Polytope (SaP) framework, which learns a set of linear half-space
constraints in the hidden state space to define a convex polytope
separating safe from unsafe representations.
SaP operates as a post-hoc method that does not modify model weights,
instead learning constraint surfaces in representation space that enable
both detection and geometric steering of unsafe outputs.
The learned polytope facets exhibit specialization in detecting different
semantic notions of safety, providing interpretable insights into the
structure of safety representations.

However, SaP's formulation leaves two questions open.
First, \emph{how many constraint planes $K$ are needed?}
\citet{chen2025sap} swept $K$ from 1 to 60 across all 14 BeaverTails
categories jointly, finding that defense performance stabilizes around
$K{=}20$--$30$ and classification accuracy plateaus near $K{=}30$--$40$.
When we instead train \emph{per-category} classifiers with random
initialization, the optimal $K$ still varies widely ($K{=}4$--$25$),
forcing expensive sweeps for each category.
Second, \emph{is the flat polytope geometry optimal?}
Recent work on concept geometry in LLMs has shown that categorical
concepts form simplices and hierarchical concepts occupy orthogonal
subspaces with branching structure~\cite{park2024lrh}---geometries
that a cone-shaped constraint may capture more naturally than flat
half-spaces.
We address both questions: SAE feature extraction largely resolves the
first, and the cone constraint explores the second.

\paragraph{Deployment model.}
\citet{chen2025sap} showed that individual polytope facets
spontaneously specialize to different harm categories even when
category labels are withheld during training.
We make this specialization explicit by training one constraint set per
harm category, which enables per-category tuning of the constraint
count and geometry.
At inference, each decoding step extracts the last-token hidden state
from a designated LLM layer and passes it through a lightweight
feature extractor (SAE encoder: a single matrix multiply plus ReLU)
followed by the constraint check.
Because our SAE-based approach reduces $K$ to 2 per category
(vs.\ $K{=}20$--$30$ in SaP), the constraint check involves a
$(B {\times} 2)$ matrix multiply instead of $(B {\times} 30)$---an
order-of-magnitude reduction in per-step cost that compounds over the
full generation sequence.
The entire check takes $<$1\,ms and can trigger output steering or
rejection.
The LLM weights are never modified: training uses pre-extracted,
frozen hidden states, and inference adds only a forward hook.
We evaluate on \emph{in-distribution BeaverTails classification}; full
adversarial robustness (e.g., against GCG~\cite{zou2023gcg} or
adaptive jailbreaks) is outside the scope of this paper and deferred
to future work.

Our contributions are:
\begin{enumerate}
\item \textbf{Post-hoc safety detection with automatic constraint
  determination via SAE}
  (\S\ref{sec:numphi}):
  With a sparse autoencoder (SAE) as feature extractor, $K{=}2$
  becomes optimal for 12/14 categories on Qwen3.5-9B and 10/14 on
  Qwen2-1.5B---largely eliminating per-category $K$ sweeps.
  On Qwen3.5-9B, SAE-initialized polytopes at $K{=}2$ achieve
  96--99\% accuracy per category
  (Tables~\ref{tab:init-vs-k-qwen2},~\ref{tab:init-vs-k-qwen35},
  Figure~\ref{fig:init-comparison}), comparable to the best accuracy
  found by exhaustive sweeps with random initialization.
  This convergence to two planes provides suggestive evidence that, in
  this safety classification setting, category boundaries admit a
  low-dimensional linear description in the SAE feature space,
  consistent with the Linear Representation Hypothesis.

\item \textbf{Interpretable, category-adaptive cone constraints}
  (\S\ref{sec:geometry}):
  For categories whose unsafe representations are angularly
  concentrated, a cone constraint---whose axis and aperture have clear
  geometric meaning for post-hoc inspection---tends to outperform the
  flat polytope (Figure~\ref{fig:geometry-comparison}).
  We introduce a three-phase training schedule
  (Algorithm~\ref{alg:threephase}; Figure~\ref{fig:three-phase},
  Table~\ref{tab:step-alloc}) that eliminates the catastrophic angle
  collapse of na\"ive training, and show that SAE initialization
  selectively benefits the cone
  (12/13 wins, $p{<}0.001$; Figure~\ref{fig:sae-synergy}) but not the
  polytope ($p{=}0.11$).
\end{enumerate}

\section{Background}\label{sec:background}

\subsection{Safety Polytope (SaP)}

The SaP framework~\cite{chen2025sap} learns $K$ linear half-space
constraints in a feature space to classify hidden states as safe or
unsafe.
Given a learnable feature extractor $g\!: \R^d \to \R^{d'}$, the safe
region is defined as:
\begin{equation}\label{eq:safe-region}
  \mathcal{S} = \bigcap_{k=1}^{K}
  \bigl\{\, \bz \in \R^{d'} \;\big|\;
  \bphi_k^\top \bz \leq \xi_k \,\bigr\},
\end{equation}
where $\bz = g(\bh)$, $\bphi_k$ are the constraint normals, and
$\xi_k$ are the thresholds.
A sample is classified as safe if and only if its representation lies
inside the polytope, i.e., $g(\bh) \in \mathcal{S}$.
The model is trained with a hinge-like loss that pushes unsafe samples
to violate at least one constraint by margin $m$, while keeping safe
samples inside with the same margin.
Entropy regularization encourages balanced usage across constraints.
Importantly, SaP is a post-hoc method: it does not modify the LLM's
weights, but learns constraint surfaces over frozen hidden states.

\subsection{Linear Representation Hypothesis and Concept Geometry}%
\label{sec:lrh}

SaP's approach rests on a broader empirical observation about how neural
networks organize information internally.
The \textbf{Linear Representation Hypothesis (LRH)} proposes that
high-level concepts are encoded as linear directions in the
representation space~\cite{park2024lrh,elhage2022superposition}.
This has been observed across diverse concept types in world
models~\cite{nanda2023othello} and confirmed by activation
interventions~\cite{li2024iti,turner2023activation,zou2023repe}:
modifying hidden states along linear directions reliably alters model
behavior in predictable ways.

However, the assumption that each concept corresponds to a single linear
direction is an oversimplification.
Recent work has shown that \emph{categorical concepts} are represented
not as single directions but as
\emph{simplices}~\cite{park2024lrh}, and that
\emph{hierarchical concepts} occupy orthogonal subspaces with branching
geometry~\cite{park2024lrh}.
These findings collectively suggest that while linear structure provides
a useful first-order approximation, safety-relevant concepts in LLMs
likely have richer geometric organization.
This motivates our exploration of constraint geometries beyond flat
half-spaces.

\section{Layer Selection}\label{sec:layer}

A fundamental design choice in applying safety constraints to hidden
states is selecting which layer to extract representations from.
Intuitively, later layers should encode more processed, semantically
rich information, but this has not been systematically verified for
safety classification.

We train polytope classifiers ($K{=}2$, random initialization) on hidden
states extracted from multiple layers of Qwen3.5-9B-Instruct,
Qwen3.5-9B-Base~\cite{qwen2025qwen3},
Qwen2-1.5B-Instruct~\cite{yang2024qwen2} (layers 4, 8, 12, 16, 20, 24,
26, 28), and gpt-oss-20b (layers 20, 24).
For each layer, we measure mean accuracy over all 14 BeaverTails
harm categories~\cite{ji2024beavertails}.

\begin{figure}[t]
  \centering
  \includegraphics[width=\columnwidth]{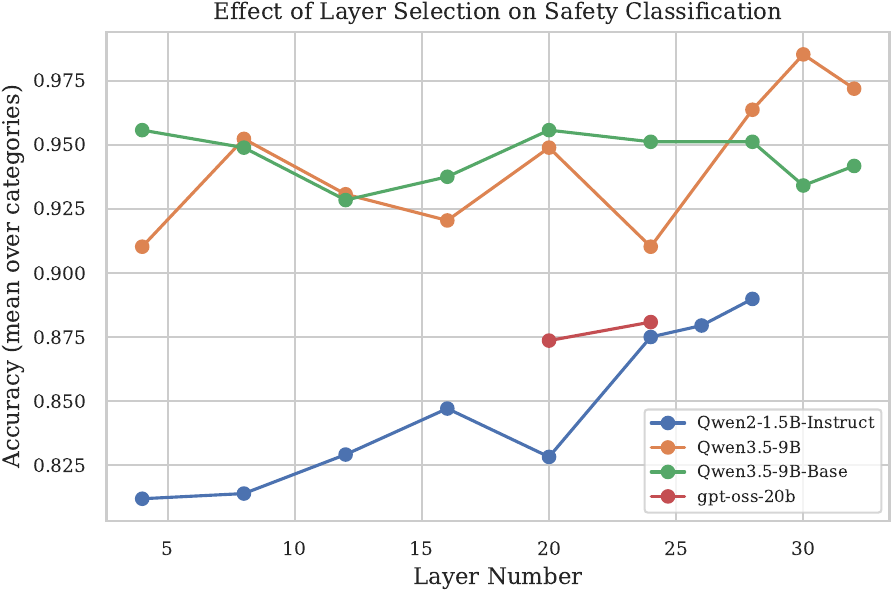}
  \caption{Mean accuracy vs.\ layer. Instruction-tuned models show
    positive trends with layer depth. Model capability dominates:
    ${\sim}8\%$ across models vs.\ ${\sim}3\%$ across layers.}
  \label{fig:layer-accuracy}
\end{figure}

Figure~\ref{fig:layer-accuracy} and Table~\ref{tab:layer-selection}
(Appendix) show two key observations.
(1)~\textbf{Model capability dominates layer choice}: the spread across
models (${\sim}8$~pp) exceeds the within-model spread across layers
(${\sim}3$~pp).
(2)~\textbf{Later layers perform better on instruction-tuned models}:
Qwen2-1.5B-Instruct shows a strong positive trend (Spearman
$\rho{=}0.929$, $p{=}0.001$; Fisher 95\% CI $[+0.62, +0.99]$), while
Qwen3.5-9B-Instruct shows a borderline trend whose confidence interval
crosses zero ($\rho{=}0.661$, $p{=}0.053$;
Fisher 95\% CI $[-0.05, +0.93]$).
The Qwen3.5-9B-Base model shows no significant trend
($\rho{=}{-}0.210$, $p{=}0.587$; Fisher 95\% CI $[-0.79, +0.56]$).

The consistent advantage of later layers on instruction-tuned models
admits two possible explanations: (1)~safety-relevant contextual
information accumulates through the forward pass, or
(2)~instruction tuning specifically concentrates safety information in
later layers.
The absence of a layer trend in the Base model supports the latter
interpretation: instruction tuning reorganizes safety representations
toward later layers.

Based on these findings, we fix \textbf{layer~32 for Qwen3.5-9B} and
\textbf{layer~28 for Qwen2-1.5B} in all subsequent experiments.
(Layer~30 achieves the highest mean accuracy on Qwen3.5-9B-Instruct
in Table~\ref{tab:layer-selection}; we chose layer~32 because it is
the final transformer block and cone experiments had already been
conducted there before the full layer sweep was completed.)
Having established where to extract representations, we next investigate
how many constraints are needed and how they should be arranged.

\section{Arranging Constraints}\label{sec:arrange}

SaP treats the number of constraint planes $K$ as a global
hyperparameter.
In this section we ask two questions: how many constraints are actually
needed (\S\ref{sec:numphi}), and whether their directions should be
initialized with geometric structure (\S\ref{sec:init}).

\subsection{How Many Constraints Are Needed?}\label{sec:numphi}

SaP requires choosing $K$, the number of constraint planes.
We sweep $K \in \{2, 3, \ldots, 7, 10, 15, 20, 25\}$ across all 14
BeaverTails categories with randomly initialized constraint normals.
For each configuration we measure accuracy, \emph{activated edges}
(the mean number of constraints violated per unsafe sample), and
\emph{effective planes} (the number of non-redundant normals after
grouping by cosine similarity ${>}\,0.95$).

\begin{figure}[t]
  \centering
  \includegraphics[width=\columnwidth]{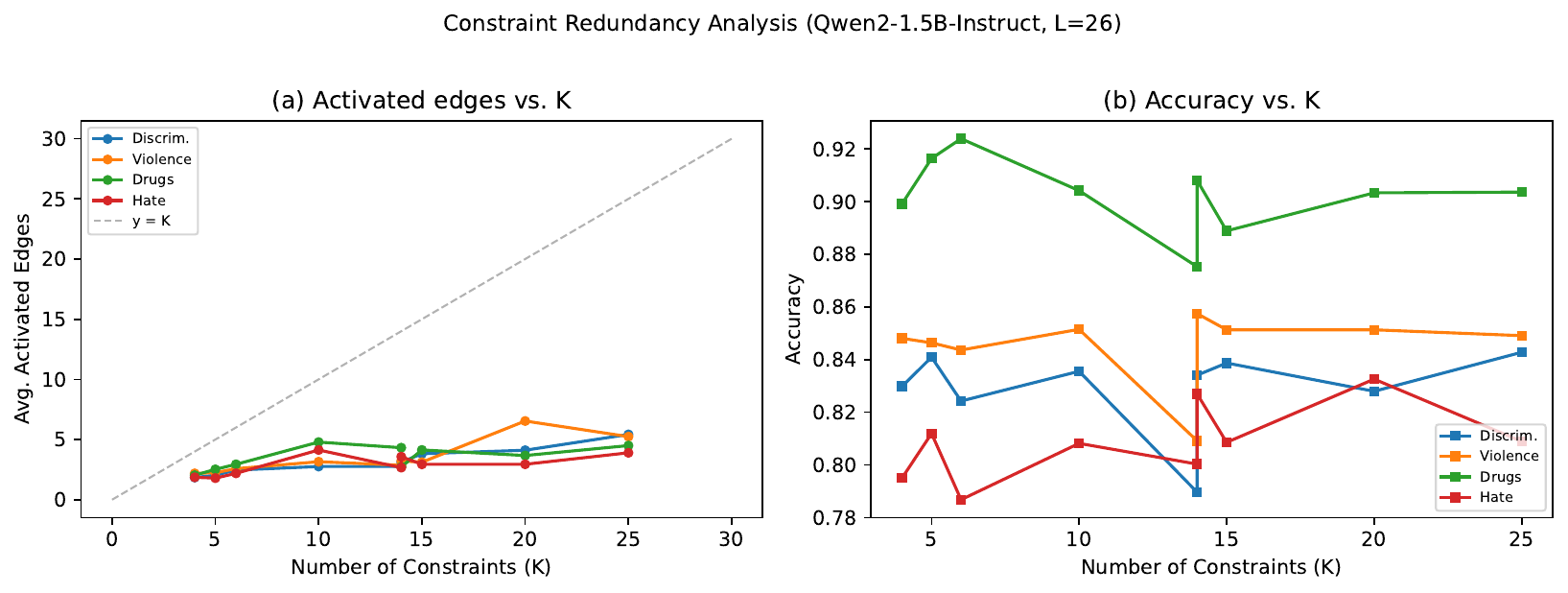}
  \caption{Constraint redundancy (Qwen2-1.5B-Instruct, L=26, random
    init). (a)~Activated edges grow proportionally with $K$ for some
    categories, indicating redundant planes. (b)~Accuracy remains flat
    despite increasing $K$.}
  \label{fig:effective-planes}
\end{figure}

The optimal $K$ varies widely: self-harm and violence favor $K{=}2$,
while hate speech and drugs prefer $K{\geq}20$.
Figure~\ref{fig:effective-planes} reveals the source of this variation.
For categories like discrimination and violence, activated edges grow
proportionally with $K$---additional planes merely duplicate existing
directions---while accuracy remains flat.
The constraints are largely redundant.

\paragraph{Automatic $K$ determination via SAE.}
To eliminate this redundancy we replace SaP's L1-only concept encoder
with a sparse autoencoder
(SAE)~\cite{cunningham2023sae,bricken2023monosemanticity}
consisting of a Linear$+$ReLU encoder and a linear decoder, trained
with MSE reconstruction loss plus L1 sparsity:
\begin{equation}\label{eq:sae-loss}
  \mathcal{L}_{\mathrm{SAE}} =
  \|\mathrm{Dec}(\mathrm{Enc}(\bh)) - \bh\|_2^2
  + \lambda_{\mathrm{sparse}}\|\mathrm{Enc}(\bh)\|_1.
\end{equation}
We first pre-train the SAE on all hidden states (10~epochs,
$d'{=}16\,384$).
Then we cluster the \emph{unsafe} samples in the SAE encoder space
using $K$-means with $K \in \{1,\ldots,10\}$, selecting the $K^*$
that maximizes the silhouette score.
The cluster centroids (normalized) serve as initial constraint normals
$\bphi_k$, and the minimum projection of each cluster's members onto
$\bphi_k$ initializes the thresholds~$\xi_k$.
During subsequent constraint training the SAE remains the feature
extractor, with a joint loss
$\mathcal{L} = \mathcal{L}_{\mathrm{constraint}}
  + w_{\mathrm{recon}}\|\mathrm{Dec}(\mathrm{Enc}(\bh)) - \bh\|_2^2
  + w_{\mathrm{sparse}}\|\mathrm{Enc}(\bh)\|_1$.

With this procedure, the picture changes dramatically:
\textbf{$K{=}2$ is optimal for 12/14 categories on both
Qwen3.5-9B-Instruct and Qwen3.5-9B-Base, and 10/14 on
Qwen2-1.5B-Instruct}
(Tables~\ref{tab:init-vs-k-qwen2} and~\ref{tab:init-vs-k-qwen35}).
The remaining categories require only $K{=}3$--$5$.

\begin{table*}[t]
\centering
\caption{Optimal $K$ by initialization method (Qwen2-1.5B-Instruct,
  best across layers 4--28). With random init the optimal $K$ ranges
  from 5 to 25; SAE reduces it to 2--5.}
\label{tab:init-vs-k-qwen2}
\small
\begin{tabular}{l cc cc cc}
\toprule
 & \multicolumn{2}{c}{Random} & \multicolumn{2}{c}{Cluster}
 & \multicolumn{2}{c}{SAE} \\
\cmidrule(lr){2-3}\cmidrule(lr){4-5}\cmidrule(lr){6-7}
Category & $K$ & Acc & $K$ & Acc & $K$ & Acc \\
\midrule
Animal   &  6 & .959 & 29 & .952 &  2 & .936 \\
Child    &  5 & .931 &  6 & .922 &  2 & .833 \\
Polit.   & 14 & .802 &  4 & .814 &  3 & .823 \\
Discr.   & 25 & .843 &  7 & .851 &  2 & .842 \\
Drugs    & 10 & .940 & 25 & .948 &  5 & .940 \\
Finan.   & 20 & .920 & 20 & .918 &  5 & .906 \\
Hate     & 25 & .854 &  4 & .841 &  2 & .828 \\
Misinf.  &  6 & .709 & 20 & .726 &  3 & .694 \\
Unethic. & 20 & .837 & 17 & .811 &  2 & .803 \\
Priv.    &  5 & .943 &  4 & .932 &  2 & .922 \\
Self-h.  & 15 & .929 &  5 & .922 &  2 & .933 \\
Sexual   & 10 & .933 &  4 & .936 &  2 & .908 \\
Terror.  &  6 & .895 & 20 & .905 &  2 & .905 \\
Viol.    & 15 & .882 & 20 & .882 &  2 & .884 \\
\midrule
$K{=}2$ count
 & \multicolumn{2}{c}{0/14}
 & \multicolumn{2}{c}{0/14}
 & \multicolumn{2}{c}{\textbf{10/14}} \\
\bottomrule
\end{tabular}
\end{table*}

\begin{figure}[t]
  \centering
  \includegraphics[width=\columnwidth]{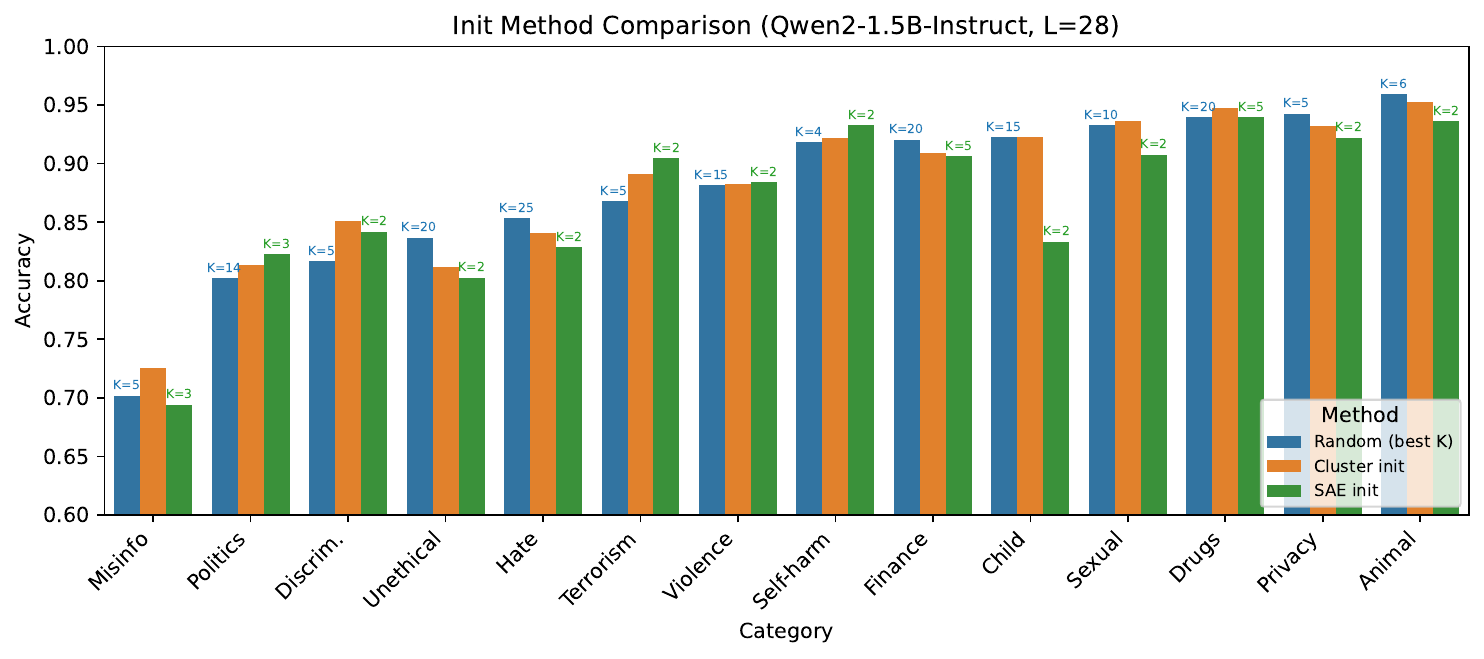}
  \caption{Accuracy comparison across initialization methods
    (Qwen2-1.5B-Instruct, L=28). Blue bars show the best accuracy found
    by exhaustive $K$ sweep with random init (annotated $K$ values).
    Green bars show SAE init at $K{=}2$--$3$. SAE achieves comparable
    accuracy \emph{without any $K$ tuning}.}
  \label{fig:init-comparison}
\end{figure}

\begin{table*}[t]
\centering
\caption{Optimal $K$ with cluster and SAE initialization on Qwen3.5-9B
  (L=32; Animal Instruct SAE from L=30, no L=32 run available).
  SAE achieves $K{=}2$ for 12 of 14 categories on each variant.}
\label{tab:init-vs-k-qwen35}
\small
\begin{tabular}{l cc cc cc cc}
\toprule
 & \multicolumn{4}{c}{Qwen3.5-9B-Instruct}
 & \multicolumn{4}{c}{Qwen3.5-9B-Base} \\
\cmidrule(lr){2-5}\cmidrule(lr){6-9}
 & \multicolumn{2}{c}{Cluster} & \multicolumn{2}{c}{SAE}
 & \multicolumn{2}{c}{Cluster} & \multicolumn{2}{c}{SAE} \\
\cmidrule(lr){2-3}\cmidrule(lr){4-5}\cmidrule(lr){6-7}\cmidrule(lr){8-9}
Category & $K$ & Acc & $K$ & Acc & $K$ & Acc & $K$ & Acc \\
\midrule
Animal   &  2 & .955 &  2 & .985 &  2 & .961 & -- &  -- \\
Child    & 29 & .922 &  2 & .957 & 30 & .943 &  2 & .966 \\
Polit.   &  2 & .961 &  2 & .959 &  3 & .961 &  2 & .967 \\
Discr.   &  3 & .965 &  2 & .962 &  8 & .964 &  2 & .966 \\
Drugs    &  2 & .984 &  2 & .990 &  2 & .985 &  2 & .987 \\
Finan.   &  3 & .975 &  2 & .980 &  2 & .979 &  2 & .980 \\
Hate     &  2 & .967 &  2 & .967 &  2 & .970 &  3 & .970 \\
Misinf.  &  3 & .938 &  2 & .949 &  3 & .943 &  2 & .940 \\
Unethic. &  4 & .959 &  3 & .963 &  3 & .813 &  2 & .806 \\
Priv.    &  2 & .984 &  2 & .984 &  2 & .944 &  2 & .934 \\
Self-h.  & 30 & .965 &  2 & .975 &  2 & .879 &  2 & .894 \\
Sexual   & -- &  -- & -- &  -- &  2 & .919 &  2 & .924 \\
Terror.  & 30 & .964 &  2 & .964 &  2 & .882 &  2 & .900 \\
Viol.    &  2 & .959 &  2 & .972 &  2 & .896 &  2 & .845 \\
\midrule
$K{=}2$ count
 & \multicolumn{2}{c}{6/14}
 & \multicolumn{2}{c}{\textbf{12/14}}
 & \multicolumn{2}{c}{9/14}
 & \multicolumn{2}{c}{\textbf{12/14}} \\
\bottomrule
\end{tabular}
\end{table*}

Random initialization places constraint normals at arbitrary
orientations, requiring many planes to ``cover'' the unsafe region
through redundant overlapping.
The SAE, by learning a sparse, geometry-preserving representation,
aligns the feature space so that the safety-relevant structure is
captured in a low-dimensional subspace.
In this aligned space, two well-oriented planes suffice to delineate
safe from unsafe regions.

This convergence to $K{=}2$ has a notable geometric implication.
In the SAE feature space, each harm category's safe--unsafe boundary
tends to be captured by just two linear directions---an observation
consistent with the Linear Representation Hypothesis
(\S\ref{sec:lrh}).
While we do not claim this as a proof of LRH, it provides suggestive
evidence that, in this BeaverTails classification setting, safety
boundaries admit a low-dimensional linear description once the feature
space is appropriately structured.

Practically, this means practitioners can fix $K{=}2$ as a default
without per-category tuning.
It also raises a natural question: if the feature space is this
well-structured, can we do even better by matching the constraint
\emph{shape} to the cluster geometry?
We address this in \S\ref{sec:geometry}.

\subsection{Direction Initialization}\label{sec:init}

The redundancy observed above suggests that constraints should be
initialized along distinct concept directions rather than randomly.
We compare three strategies for $\bphi_k$:
\textbf{(1)~Random}: normalized Gaussian vectors.
\textbf{(2)~Cluster}: we project unsafe hidden states into a
50-dimensional PCA space, run $K$-means with $K \in \{2, \ldots, 30\}$,
select the $K^*$ maximizing the silhouette score, and use the normalized
cluster centroids (in the original feature space) as initial normals.
\textbf{(3)~SAE}: we pre-train the SAE (Eq.~\ref{eq:sae-loss}), cluster
in the SAE encoder space with $K \in \{1, \ldots, 10\}$, and use the
resulting centroids as normals.
On Qwen2-1.5B, structured initialization (cluster and SAE) at $K{=}2$
achieves accuracy comparable to the best random-init result found by
sweeping $K{=}3$--$25$ (Figure~\ref{fig:init-comparison})---but
\emph{without requiring any $K$ search}.
Building on this, we applied cluster and SAE initialization to
Qwen3.5-9B (L=32), where random-init data is unavailable.
The results (Table~\ref{tab:init-vs-k-qwen35}) show consistently high
accuracy: SAE at $K{=}2$ achieves $.96$--$.99$ across 13 categories,
and cluster initialization produces comparable results.
This confirms that structured initialization is not merely a shortcut
to match random-init performance, but enables strong classification on
larger models where exhaustive $K$ sweeps were never conducted.

The polytope's tolerance to initialization direction (given sufficient
$K$) reflects the fact that flat half-space constraints lack a
distinguished axis: gradient descent can rotate $\bphi_k$ to any
orientation.
This changes dramatically for the \emph{cone}, where the axis direction
carries geometric significance---as we show in \S\ref{sec:sae_synergy}.

\section{Geometric Extensions}\label{sec:geometry}

The flat half-space constraints of the polytope assume that the boundary
between safe and unsafe regions can be well-approximated by hyperplane
intersections.
However, if unsafe concepts radiate from common origins in
representation space, a non-flat constraint surface may be more
appropriate.

\subsection{Cone Constraint}\label{sec:cone}

If unsafe representations concentrate along specific directions rather
than filling a half-space, a cone-shaped boundary may fit them more
tightly.
We define:

\begin{definition}[Cone constraint]\label{def:cone}
Each constraint $k$ has unit axis
$\hat{\bphi}_k = \bphi_k/\|\bphi_k\|$, apex
$\mathbf{a}_k = \xi_k \hat{\bphi}_k$, and learnable aperture
$\alpha_k \in [\alpha_{\min}, \alpha_{\max}]$.
The constraint score is
\begin{equation}\label{eq:cone-score}
  s_k^{\mathrm{cone}}(\bz) =
  (\bz - \mathbf{a}_k)^{\!\top} \hat{\bphi}_k
  - \|\bz - \mathbf{a}_k\|\,\cos\alpha_k\,.
\end{equation}
A point $\bz = g(\bh)$ is unsafe if $s_k^{\mathrm{cone}} > 0$ for
any $k$, i.e.,
$\angle(\bz{-}\mathbf{a}_k,\,\hat{\bphi}_k) < \alpha_k$.
The cone encloses the unsafe region around the axis; points outside
all cones are safe.
When $\alpha_k = \pi/2$ the cosine term vanishes and the score reduces
to the polytope score $\hat{\bphi}_k^{\!\top}\bz - \xi_k$.
\end{definition}

\begin{definition}[Cylinder constraint]\label{def:cylinder}
Each constraint $k$ has unit axis $\hat{\bphi}_k$, axial extent
$[0, \xi_k]$, and learnable radius $r_k$.
Let $a_k = \bz^{\!\top}\hat{\bphi}_k$ be the axial projection and
$d_k^{\perp} = \|\bz - a_k\hat{\bphi}_k\|$ the perpendicular distance.
The constraint score is
\begin{equation}\label{eq:cyl-score}
  s_k^{\mathrm{cyl}}(\bz) =
  \max\!\bigl(\,r_k - d_k^{\perp},\;
  -a_k,\;
  a_k - \xi_k\,\bigr).
\end{equation}
A point is safe (score $\leq 0$) iff it lies outside the cylinder
radially ($d_k^{\perp} \geq r_k$) \emph{and} within the axial range
($0 \leq a_k \leq \xi_k$) for all~$k$.
\end{definition}

\subsubsection{Three-Phase Training}\label{sec:threephase}

Na\"ive joint training of the cone suffers from \emph{catastrophic angle
collapse}: the aperture saturates to $\alpha_{\max}{=}90^{\circ}$,
reducing the cone to a degenerate half-space.
To prevent this, we introduce a three-phase schedule
(Algorithm~\ref{alg:threephase}):
(1)~\textbf{Axis phase}: train $\bphi_k, \xi_k$ with $\alpha_k$ frozen;
(2)~\textbf{Angle phase}: freeze axis, train $\alpha_k$ with shrink
ratio $\gamma$;
(3)~\textbf{Joint phase}: train all parameters jointly.

\begin{algorithm}[t]
\caption{Three-Phase Cone Training}\label{alg:threephase}
\KwInput{Hidden states $\{(\bh_i, y_i)\}$, encoder $g$}
\KwHyper{$e_a, e_\alpha, e_j$: epoch counts;\;
  $\gamma$: shrink ratio;\;
  $\alpha_{\mathrm{init}}$: initial aperture}
\KwOutput{Trained $\bphi_k, \xi_k, \alpha_k$}
\BlankLine
Initialize $\bphi_k, \xi_k$; set $\alpha_k \leftarrow \alpha_{\mathrm{init}}$\;
\textbf{Phase 1} (Axis): \For{$e_a$ epochs}{
  Update $\bphi_k, \xi_k$ with $\alpha_k$ frozen\;
}
$\alpha_k \leftarrow \gamma \cdot \alpha_k$ \tcp*{shrink}
\textbf{Phase 2} (Angle): \For{$e_\alpha$ epochs}{
  Update $\alpha_k$ with $\bphi_k, \xi_k$ frozen\;
}
\textbf{Phase 3} (Joint): \For{$e_j$ epochs}{
  Update $\bphi_k, \xi_k, \alpha_k$ jointly\;
}
\end{algorithm}

\begin{figure}[t]
  \centering
  \includegraphics[width=\columnwidth]{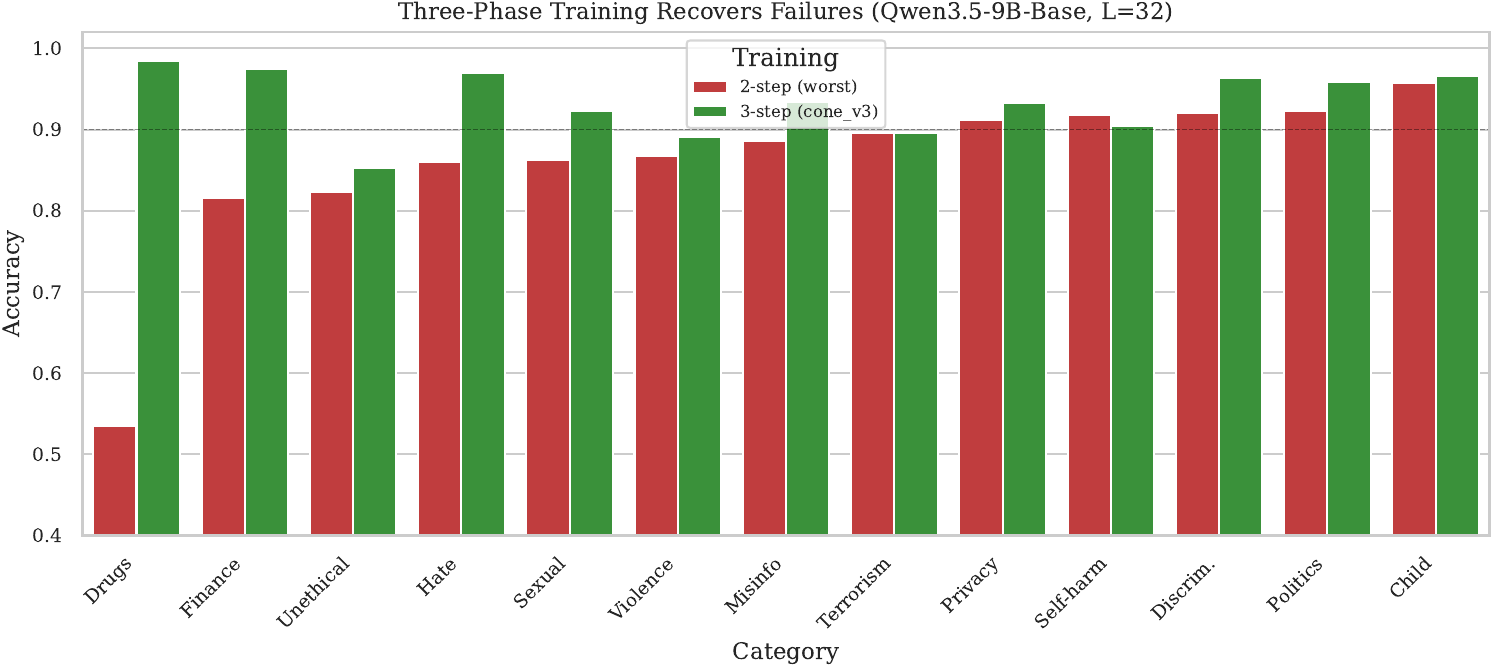}
  \caption{Three-phase training eliminates catastrophic failures
    (Qwen3.5-9B-Base, L=32). Red: worst 2-step accuracy per category.
    Green: best 3-step. Dashed line: 0.90 accuracy threshold.}
  \label{fig:three-phase}
\end{figure}

On Qwen3.5-9B-Base (L=32), the worst 2-step accuracy is $0.535$ (drug
abuse), compared to ${\geq}0.93$ for all categories with the 3-step
schedule (Figure~\ref{fig:three-phase}).
The best configuration ($e_a{=}18$, $e_\alpha{=}1$, $e_j{=}3$,
$\gamma{=}0.7$, $\alpha_{\mathrm{init}}{=}80^{\circ}$) achieves mean
accuracy $0.967$ on Qwen3.5-9B-Instruct (Table~\ref{tab:step-alloc}).

\begin{table}[t]
\centering
\caption{Top cone step allocations (Qwen3.5-9B, L=32, $K{=}2$).}
\label{tab:step-alloc}
\small
\begin{tabular}{rrrrrccc}
\toprule
$e_a$ & $e_\alpha$ & $e_j$ & $\gamma$ & $\alpha_0$ & Acc & FPR & Angle \\
\midrule
18 & 1 & 3 & 0.7 & 80 & \textbf{.967} & .033 & 87.1 \\
10 & 1 & 10 & 0.7 & 50 & .964 & .029 & 74.1 \\
15 & 1 & 5 & 0.7 & 50 & .961 & .026 & 74.9 \\
10 & 1 & 10 & 0.7 & 80 & .963 & .030 & 88.0 \\
20 & 1 & 5 & 0.7 & 60 & .958 & .028 & 77.2 \\
\bottomrule
\end{tabular}
\end{table}

The collapse occurs because axis direction and aperture angle interact:
a poorly oriented axis causes the optimizer to widen the aperture, but a
wide aperture provides weak gradient signal for axis correction---a
vicious cycle converging to $\alpha{=}90^{\circ}$.
The three-phase schedule breaks this by stabilizing the axis first, then
narrowing the aperture with the shrink ratio $\gamma$ as a warm-start.

\subsubsection{SAE--Cone Synergy}\label{sec:sae_synergy}

\begin{figure}[t]
  \centering
  \includegraphics[width=\columnwidth]{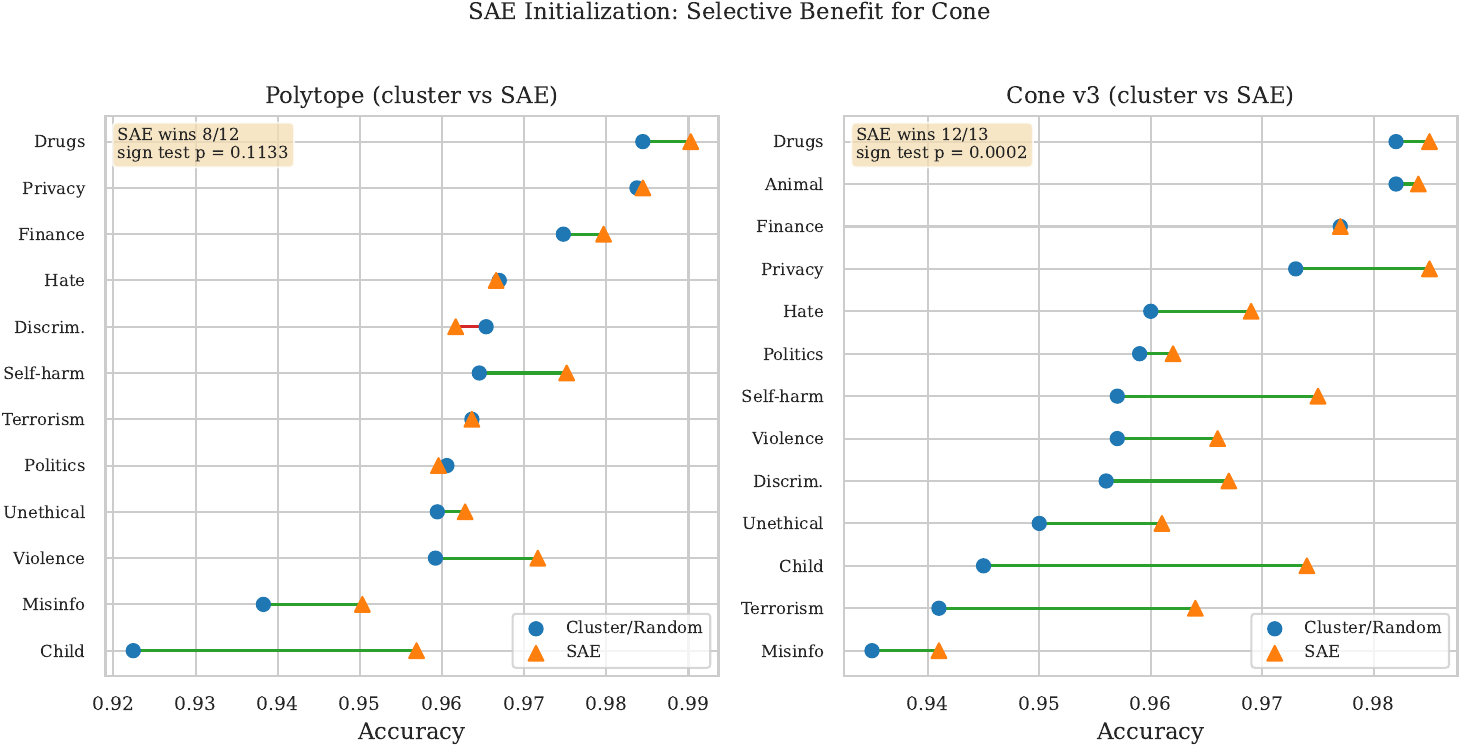}
  \caption{SAE init selectively benefits the cone (Qwen3.5-9B-Instruct,
    L=32). Left: \poly{} ($p{=}0.11$, not significant). Right: \cone{}
    (12/13 SAE wins, $p{<}0.001$).}
  \label{fig:sae-synergy}
\end{figure}

In \S\ref{sec:init}, initialization direction did not matter for the
polytope.
We now test whether this holds for the cone, whose axis has geometric
significance.
We compare SAE-based vs.\ cluster-based initialization separately for
polytope and cone (Qwen3.5-9B, L=32), using paired comparison across
13 categories.
Figure~\ref{fig:sae-synergy} shows a striking asymmetry:
the \poly{} shows no significant benefit from SAE initialization
(SAE wins 8/12 non-tie categories; binomial $p{=}0.11$ one-sided,
Wilson 95\% CI on win rate $[0.43, 0.90]$),
while the \cone{} benefits strongly (SAE wins 12/13 non-tie categories;
binomial $p{<}0.001$, Wilson 95\% CI $[0.76, 1.00]$).
A Benjamini--Hochberg correction applied across the 14 per-category
comparisons preserves the cone's effect at $q{<}0.01$.

The polytope's $\bphi_k$ defines a hyperplane with no distinguished
axis---gradient descent can rotate it freely regardless of
initialization.
The cone, by contrast, has an axis that determines the direction around
which the aperture is measured: a well-chosen axis lets the cone tightly
enclose the angular spread of the unsafe cluster.
The SAE identifies sparse feature directions along which unsafe samples
concentrate, providing an axis that gradient descent preserves.
This result suggests that the SAE-initialized cone captures unsafe
cluster geometry more accurately than the flat polytope, connecting our
two main contributions: the SAE not only reduces $K$
(\S\ref{sec:numphi}) but also provides geometrically meaningful
directions that enable non-flat shapes to exploit angular cluster
structure.

\subsubsection{Geometric Analysis of the Cone}\label{sec:kappa}

Having established \emph{how} to train cones (\S\ref{sec:threephase})
and initialize them (\S\ref{sec:sae_synergy}), we examine \emph{when}
the cone outperforms the polytope and what geometry the cone learns.

\begin{figure}[t]
  \centering
  \includegraphics[width=\columnwidth]{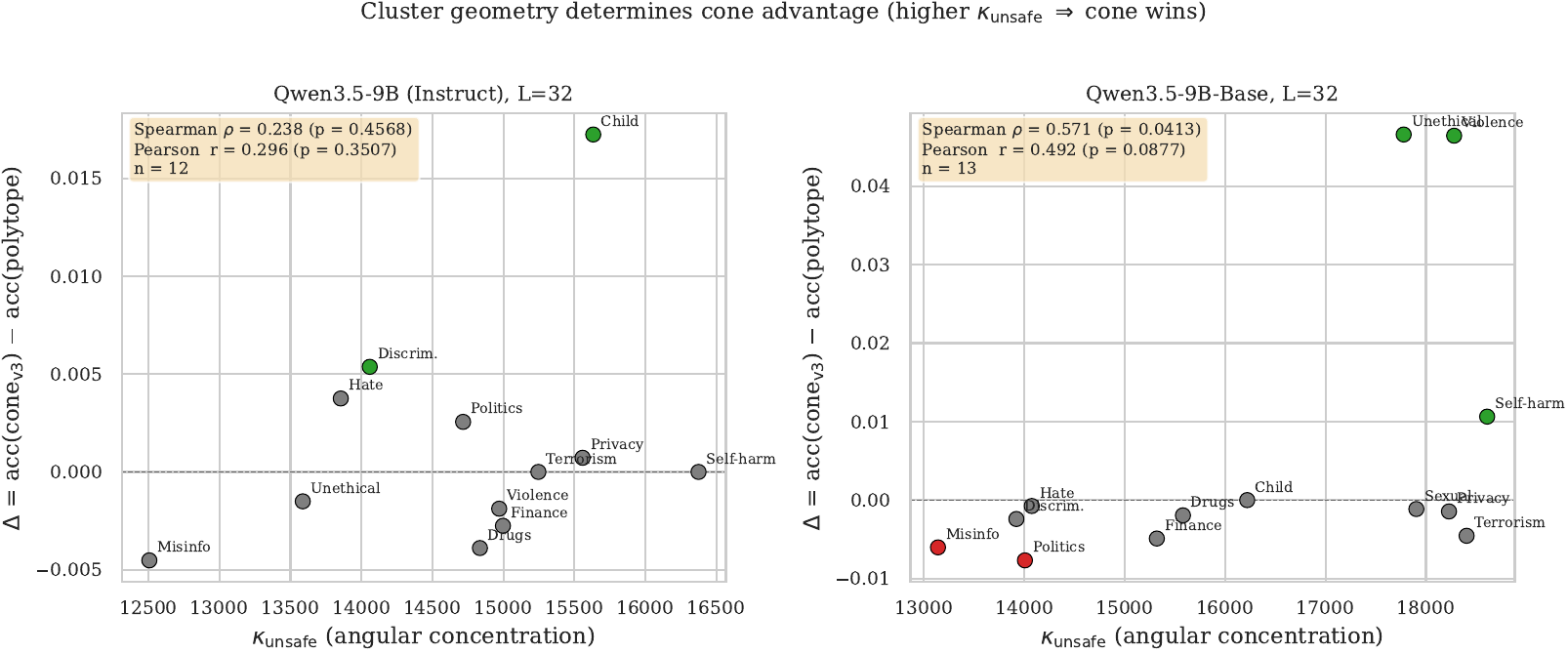}
  \caption{$\kunsafe$ vs.\ $\Delta(\cone{-}\poly)$ at L=32,
    apples-to-apples SAE-initialized pairing.
    Higher concentration is associated with cone advantage:
    Instruct $\rho{=}{+}0.238$ ($p{=}0.46$, $n{=}12$);
    Base $\rho{=}{+}0.571$ ($p{=}0.041$, $n{=}13$).
    Green: cone wins; red: polytope wins; gray: tie.}
  \label{fig:kappa-vs-delta}
\end{figure}

For each of the 14 categories we compute the von Mises--Fisher
concentration $\kunsafe$~\cite{banerjee2005vmf} of the unsafe cluster
at L=32 and the accuracy gap
$\Delta = \mathrm{acc}(\cone) - \mathrm{acc}(\poly)$,
using best-accuracy SAE-initialized constraints per category.
On Qwen3.5-9B-Base the correlation is moderately positive:
Spearman $\rho{=}{+}0.571$, $p{=}0.041$
(Fisher 95\% CI $[-0.01, +0.86]$, $n{=}13$).
On Qwen3.5-9B-Instruct the point estimate is still positive but its
confidence interval crosses zero: $\rho{=}{+}0.238$, $p{=}0.46$
(Fisher 95\% CI $[-0.42, +0.73]$, $n{=}12$;
Figure~\ref{fig:kappa-vs-delta}).
The highest-$\kappa$ categories on Instruct
(child abuse: $\kappa{=}15\,633$, $\Delta{=}{+}0.017$) and on Base
(non-violent unethical: $\kappa{=}17\,779$, $\Delta{=}{+}0.047$;
violence: $\kappa{=}18\,281$, $\Delta{=}{+}0.046$) benefit from the
cone, while low-$\kappa$ categories show no consistent advantage.
We treat this as a preliminary observation: the sign is consistent
across both model variants, but statistical power at $n{=}13$ is
insufficient to claim a significant effect on the Instruct variant.

\begin{figure}[t]
  \centering
  \includegraphics[width=\columnwidth]{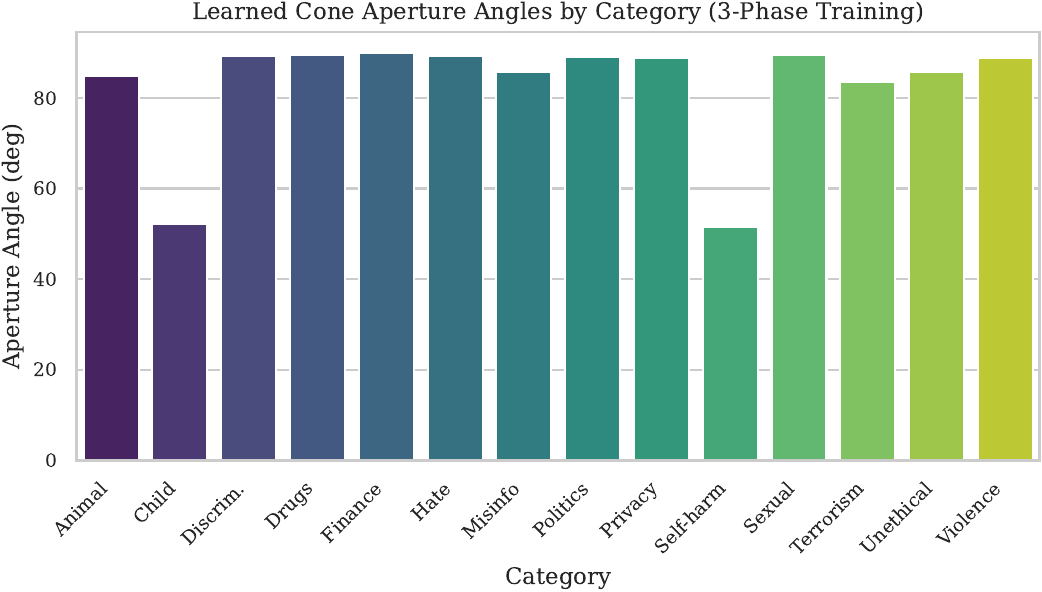}
  \caption{Learned aperture angles (3-phase, Qwen3.5-9B-Instruct, L=32).
    High-$\kunsafe$ categories learn tighter angles; diffuse categories
    approach $90^{\circ}$.}
  \label{fig:cone-angles}
\end{figure}

The learned aperture angles (Figure~\ref{fig:cone-angles}) corroborate
this trend: some categories (e.g., child abuse, self-harm) learn
noticeably tighter angles (${\sim}50^{\circ}$), while most diffuse
categories converge to $70$--$90^{\circ}$, near the polytope limit
($\alpha{=}90^{\circ}$).
The cone adaptively interpolates between a tight angular boundary and a
flat half-space depending on the cluster's angular concentration.

\subsection{Cylinder --- A Negative Result}\label{sec:cylinder}

To test whether \emph{any} geometric extension helps, we implement a
cylinder constraint (Definition~\ref{def:cylinder}).

\begin{figure}[t]
  \centering
  \includegraphics[width=\columnwidth]{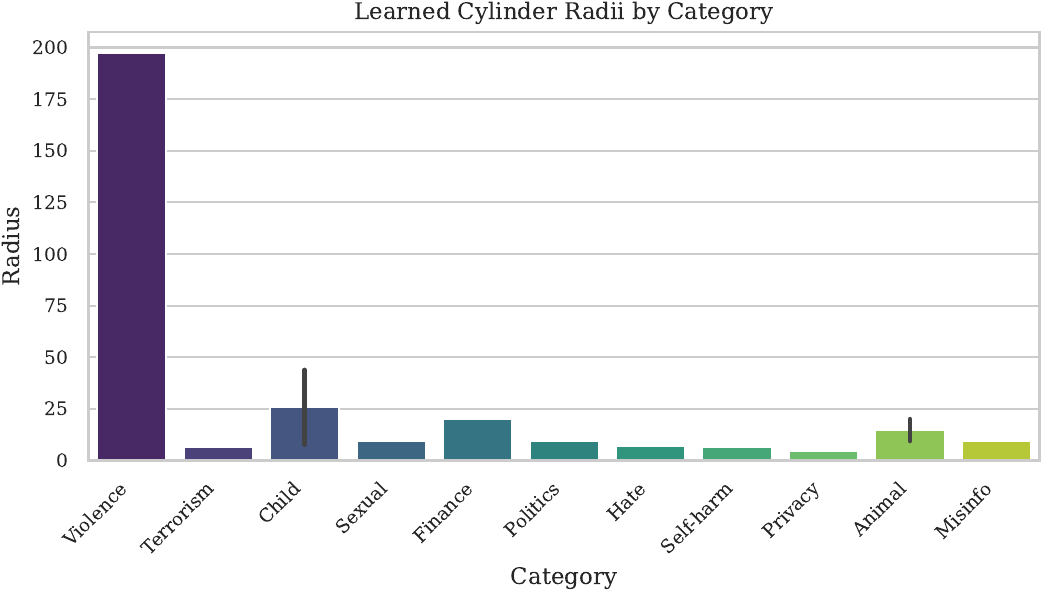}
  \caption{Cylinder radii by category (Qwen2-1.5B, L=26). Does not
    improve over \poly{}.}
  \label{fig:cylinder-radius}
\end{figure}

On Qwen2-1.5B, the cylinder does not exceed polytope accuracy for any
category (Table~\ref{tab:ablation-qwen2} in Appendix~\ref{app:ablation};
learned radii shown in Figure~\ref{fig:cylinder-radius}).

This establishes that the ``add a geometric parameter'' recipe is not
universally effective.
The cone's advantage is specific to its match with \emph{angular}
cluster structure (\S\ref{sec:kappa}).
The cylinder constrains \emph{radial} extent, which does not align with
the observed cluster geometry.
This reinforces that geometry-guided design requires understanding
\emph{which} geometric property is informative.

The full geometry $\times$ initialization ablation is provided in
Table~\ref{tab:ablation} and Figure~\ref{fig:geometry-comparison}
(Appendix~\ref{app:ablation}).

\subsection{Unified All-Categories Training}\label{sec:all-mixed}

Per-category training (\S\ref{sec:cone}, \S\ref{sec:cylinder}) fits a
separate constraint model per harm category.
To verify that the geometric advantage holds without
category-specialization, we additionally train a single
\emph{unified} model on the union of all 14 BeaverTails unsafe
categories balanced against the safe category, on Qwen3.5-9B hidden
states from layer 32 (268{,}370 train + 29{,}414 test samples).
The total number of constraints is fixed to
$K_{\text{tot}}{=}\sum_c K^*_c{=}36$, the sum of per-category $K^*$
values selected by the SAE initialization in \S\ref{sec:init}.

\begin{table}[t]
  \centering
  \small
  \caption{In-distribution test metrics on the unified
    all-categories balanced split (Qwen3.5-9B, $L{=}32$,
    $K{=}36$). \cone{} achieves the highest accuracy and F1; \poly{}
    is consistently lowest because its $\phi$ vectors are not
    norm-normalized. Per-category breakdown:
    Appendix~\ref{app:all-mixed-percat}.}
  \label{tab:all-mixed}
  \begin{tabular}{lcccc}
    \toprule
    Geometry & ACC & F1 & FPR & FNR \\
    \midrule
    \poly{}                  & 0.940 & 0.941 & 0.068 & \textbf{0.052} \\
    \cyl{}                   & 0.950 & 0.948 & \textbf{0.017} & 0.083 \\
    three-phase \cone{}      & 0.953 & 0.952 & 0.026 & 0.068 \\
    \cone{}                  & \textbf{0.956} & \textbf{0.955} & 0.029 & 0.059 \\
    \bottomrule
  \end{tabular}
\end{table}

\paragraph{\cone{} consistently wins.}
The unified \cone{} achieves the highest accuracy (0.956) and F1
(0.955), matching the per-category trend
(Table~\ref{tab:ablation}).
Evaluating the same unified models on each of the 14 per-category
test splits (Appendix~\ref{app:all-mixed-percat}) shows \cone{} wins
on $9/14$ categories and \poly{} wins on $0/14$.

\paragraph{\cyl{} degenerates radially.}
Of the 36 trained cylinder radii, 27 collapse to exactly $0$ and only
one retains a meaningful value ($16.5$ feature-space units against
the upper bound $200$).
With the radial constraint silenced, cylinder discrimination relies
entirely on the axial bound $0 \le \phi^\top h \le \theta$, reducing
the cylinder to an axial-slab classifier.
This category-agnostic degeneration corroborates the per-category
negative result in \S\ref{sec:cylinder}: constraining radial extent
does not help even at scale.

\paragraph{Cone half-angles partially saturate.}
$9/36$ of $\alpha_k$ values reach the upper bound $\alpha_k{=}90^\circ$
(cone reduces to half-space) while the remaining $27$ stay strictly
below $90^\circ$, with $25$ of them below $78^\circ$.
The three-phase variant ($\alpha_k$-epoch budget $=5$) leaves
$11/36$ angles fixed exactly at the initialization
$\alpha_k{=}80^\circ$, suggesting the $\alpha$-phase is too short to
fit the unified $\alpha$ landscape---hence its slightly lower
accuracy than single-phase \cone{}.

\paragraph{\poly{} underperforms due to scale heterogeneity.}
Without normalization, per-cluster $\phi$-norms range from $4.9$ to
$17.3$, so the constraints have unequal effective scale.
\cone{} and \cyl{} normalize $\phi$ internally, removing this
heterogeneity.
The resulting accuracy gap averages $1.5$--$1.6\%$ but ranges from
$1.2\%$ to $7.0\%$ across categories, with the largest gap on the
\textit{controversial topics} category
(Appendix~\ref{app:all-mixed-percat}).

\subsection{Robustness and Utility}\label{sec:robustness}

Since \cone{} reduces to \poly{} at $\alpha_k{=}\pi/2$, SaP's
robustness and utility guarantees~\cite{chen2025sap} are inherited at
the boundary.
Our in-distribution results show that cone accuracy matches or
exceeds polytope accuracy for most categories on Qwen3.5-9B
(Table~\ref{tab:ablation}, Figure~\ref{fig:geometry-comparison} in
Appendix~\ref{app:ablation}).
\citet{chen2025sap} demonstrated that the polytope preserves model
utility (MMLU) and withstands adversarial attacks (HarmBench); we did
not repeat these evaluations in this work, but extending them to cone
and cylinder constraints is a natural direction for future work.

\paragraph{External classifier reference.}
As a reference point for hidden-state classification, we ran NVIDIA
Llama-3.1-Nemotron-Safety-Guard-8B
(``NemoGuard''~\cite{rebedea2023nemoguard}) applied as either input
or output rails.

\begin{table}[t]
\centering
\caption{Moderation benchmark (Qwen2-1.5B, 482 prompts). Harmful
  block rate (higher = safer) and helpful allow rate (higher = less
  over-blocking). Polytope check latency excludes LLM forward pass
  (shared with normal inference).}
\label{tab:moderation}
\small
\begin{tabular}{lccc}
\toprule
Method & Harmful block & Helpful allow & Check (ms) \\
\midrule
Raw (no defense)     & 82.4\% & 96.5\% & --- \\
\poly{} (ours)       & \textbf{84.1\%} & 89.3\% & 0.18 \\
NemoGuard (output)   & 83.0\% & \textbf{95.5\%} & 59.5 \\
NemoGuard (input)    & 82.4\% & 89.4\% & 59.4 \\
\bottomrule
\end{tabular}
\end{table}

The \poly{} achieves the highest harmful-block rate (84.1\%) at
${\sim}330\times$ lower check latency than NemoGuard
(0.18\,ms vs.\ 59\,ms).
Its helpful-allow rate (89.3\%) is lower than NemoGuard output rails
(95.5\%), indicating more aggressive blocking of benign content---a
trade-off that may be acceptable in high-risk deployment settings.
The cone adds only one dot product and one cosine term per constraint,
so its forward cost remains sub-millisecond; a full cone evaluation on
this benchmark is deferred to future work.
Note that NemoGuard processes raw text end-to-end while our method
operates on hidden states during generation; the latency comparison
reflects \emph{classifier-only} cost (the LLM forward pass is shared
with normal inference).

\section{Related Work}\label{sec:related}

\paragraph{LLM safety and alignment.}
RLHF~\cite{ouyang2022instructgpt} and supervised
fine-tuning~\cite{bai2022constitutional} modify model weights to reduce
harmful outputs.
BeaverTails~\cite{ji2024beavertails} and
HarmBench~\cite{mazeika2024harmbench} standardize safety evaluation.
Red teaming~\cite{zou2023gcg,chao2023pair,wei2024jailbroken}
demonstrates the fragility of these defenses.
Our approach operates on internal representations rather than modifying
weights, enabling interpretable post-hoc safety mechanisms.

\paragraph{Representation engineering and activation steering.}
The discovery that safety-relevant information is linearly encoded has
enabled inference-time intervention methods.
RepE~\cite{zou2023repe} provides a framework for representation
manipulation.
Activation Addition~\cite{turner2023activation} and
ITI~\cite{li2024iti} steer behavior by modifying activation vectors.
Our multi-constraint approach explicitly models the multi-dimensional
structure of safety.

\paragraph{Sparse autoencoders and interpretability.}
SAEs~\cite{cunningham2023sae,bricken2023monosemanticity} decompose
neural activations into interpretable sparse features, addressing the
superposition hypothesis~\cite{elhage2022superposition}.
We leverage SAEs as geometry-preserving feature extractors that produce
sparse representations in which safety clusters are easily identified.

\paragraph{Geometry of concepts in LLMs.}
\citet{park2024lrh} showed that categorical concepts form simplices and
hierarchical concepts occupy orthogonal subspaces---directly relevant to
our polytope and cone constraints.
The vMF distribution~\cite{banerjee2005vmf} models directional data on
the hypersphere; we use $\kappa$ as a cluster compactness measure.

\paragraph{Safety classifiers.}
LlamaGuard~\cite{inan2023llamaguard},
NeMo~Guardrails~\cite{rebedea2023nemoguard},
WildGuard~\cite{han2024wildguard}, and OpenAI
moderation~\cite{markov2023openai} operate at the text level.
Our approach operates on hidden states for complementary, lower-latency
protection.

\section{Discussion}\label{sec:discussion}

\paragraph{Summary.}
This work demonstrates that reasoning about constraint geometry yields
practical benefits:
(1)~SAE feature extraction reduces optimal $K$ to 2 for 12/14
categories on Qwen3.5-9B and 10/14 on Qwen2-1.5B, largely eliminating
per-category hyperparameter search
(Tables~\ref{tab:init-vs-k-qwen2},~\ref{tab:init-vs-k-qwen35};
\S\ref{sec:numphi});
(2)~a three-phase training schedule stabilizes cone learning
(Figure~\ref{fig:three-phase}, Table~\ref{tab:step-alloc};
\S\ref{sec:threephase});
(3)~the cone adaptively interpolates between a tight angular boundary
and a flat half-space depending on each category's cluster
concentration
(Figures~\ref{fig:kappa-vs-delta},~\ref{fig:cone-angles};
\S\ref{sec:kappa});
(4)~SAE initialization selectively benefits the cone but not the
polytope (Figure~\ref{fig:sae-synergy}; \S\ref{sec:sae_synergy}).

\paragraph{Interpretability implications.}
Each constraint has clear geometric meaning: direction ($\bphi_k$),
threshold ($\xi_k$), and for cones, angular spread ($\alpha_k$).
This enables practitioners to inspect ``directions of harm,'' diagnose
failures by examining constraint activation, and adjust safety
boundaries post-hoc without retraining.

\paragraph{Scope and limitations.}
This paper is primarily an \emph{in-distribution classification} study
on the English-only BeaverTails dataset with per-category binary
classifiers.
Several important dimensions remain open:
(1)~\textbf{Adversarial robustness}: we do not evaluate against
adaptive jailbreaks (GCG, PAIR) or distribution-shifted prompts;
HarmBench evaluation of the cone is deferred to future work.
(2)~\textbf{Utility preservation}: MMLU and other capability
benchmarks have not been run with the cone; SaP's existing MMLU
results~\cite{chen2025sap} serve as a reference but are not directly
comparable.
(3)~\textbf{Unknown-category routing}: our classifiers are trained
per-category; detecting novel harm types not in BeaverTails is future
work.
(4)~\textbf{Multi-lingual}: all experiments are English-only.
(5)~On Qwen2-1.5B, SAE at $K{=}2$ occasionally trades accuracy for
simplicity (e.g., Child: $.833$ vs.\ $.931$ with random at $K{=}5$);
on Qwen3.5-9B the trade-off is negligible.
(6)~Several cells in the ablation tables
(Tables~\ref{tab:ablation},~\ref{tab:ablation-qwen2}) are unreported
where experimental runs were unavailable at submission time.
(7)~Cylinder tested on Qwen2-1.5B only.

\paragraph{Future work.}
\textbf{Automatic routing}: use $\kunsafe$ to select \poly{} vs.\
\cone{} per category automatically.
\textbf{Multi-lingual}: test $\kappa$--$\Delta$ transfer across
languages.
\textbf{Adversarial evaluation}: GCG~\cite{zou2023gcg} and PAIR
attacks.
\textbf{Interpretability}: connect cone axis to SAE features for
safety decision attribution.

This work used the BeaverTails dataset~\cite{ji2024beavertails} and
builds on the Safety Polytope framework~\cite{chen2025sap}.

\bibliographystyle{plainnat}
\bibliography{references}

\appendix

\section{Notation}\label{app:notation}

\begin{table}[H]
\centering
\small
\caption{Summary of notation.}
\begin{tabular}{cl}
\toprule
Symbol & Description \\
\midrule
$\bh \in \R^d$ & Hidden-state vector from LLM layer \\
$g(\cdot)$ & Feature extractor / concept encoder (SAE) \\
$\bz = g(\bh)$ & Encoded feature vector \\
$\bphi_k$ & Constraint normal / cone axis ($k$-th) \\
$\xi_k$ & Constraint threshold \\
$\alpha_k$ & Cone aperture angle \\
$r_k$ & Cylinder radius \\
$K$ & Number of constraints \\
$K^{*}$ & Auto-determined cluster count \\
$\kunsafe$ & vMF concentration of unsafe cluster \\
$\Delta$ & $\mathrm{acc}(\cone) - \mathrm{acc}(\poly)$ \\
$\gamma$ & Angle shrink ratio (3-phase training) \\
\bottomrule
\end{tabular}
\end{table}

\section{Full Ablation}\label{app:ablation}

\begin{table}[H]
\centering
\caption{Qwen3.5-9B-Instruct (L=32): \poly{} and \cone{} (3-phase)
  accuracy per category by initialization method.
  Cone cluster data from GH step summary scrape.
  ``--'' = no data.}
\label{tab:ablation}
\small
\begin{tabular}{lcccc}
\toprule
 & \multicolumn{2}{c}{\poly} & \multicolumn{2}{c}{\cone{} (3-phase)} \\
 \cmidrule(lr){2-3}\cmidrule(lr){4-5}
 & clust & SAE & clust & SAE \\
\midrule
Animal  & .955 & --   & .982 & .982 \\
Child   & .922 & .957 & .945 & .974 \\
Polit.  & .961 & .960 & .959 & .962 \\
Discr.  & .965 & .962 & .956 & .967 \\
Drugs   & .984 & .990 & .982 & .986 \\
Finan.  & .975 & .980 & .977 & .977 \\
Hate    & .967 & .967 & .960 & .970 \\
Misinf. & .938 & .950 & .935 & .946 \\
Uneth.  & .959 & .963 & .950 & .961 \\
Priv.   & .984 & .984 & .973 & .985 \\
Self-h. & .965 & .975 & .957 & .975 \\
Sexual  &   -- &   -- &   -- &   -- \\
Terror. & .964 & .964 & .941 & .964 \\
Viol.   & .959 & .972 & .957 & .970 \\
\bottomrule
\end{tabular}
\end{table}

\begin{table}[H]
\centering
\caption{Qwen2-1.5B-Instruct (L=28): \poly{} accuracy by init method,
  and \cyl{} (L=26). ``--'' = no data.}
\label{tab:ablation-qwen2}
\small
\begin{tabular}{lcccc}
\toprule
 & \multicolumn{3}{c}{\poly} & \cyl \\
 \cmidrule(lr){2-4}\cmidrule(lr){5-5}
 & rand & clust & SAE & SAE \\
\midrule
Animal  & .959 & .952 & .936 & .900 \\
Child   & .922 & .922 & .833 & .888 \\
Polit.  & .802 & .814 & .823 & .813 \\
Discr.  & .817 & .851 & .842 &   -- \\
Drugs   & .939 & .948 & .940 &   -- \\
Finan.  & .920 & .909 & .906 & .854 \\
Hate    & .854 & .841 & .828 & .786 \\
Misinf. & .702 & .726 & .694 & .622 \\
Uneth.  & .837 & .811 & .803 &   -- \\
Priv.   & .943 & .932 & .922 & .920 \\
Self-h. & .918 & .922 & .933 & .922 \\
Sexual  & .933 & .936 & .908 & .869 \\
Terror. & .868 & .891 & .905 & .859 \\
Viol.   & .882 & .882 & .884 & .606 \\
\bottomrule
\end{tabular}
\end{table}

\begin{figure}[H]
  \centering
  \includegraphics[width=\columnwidth]{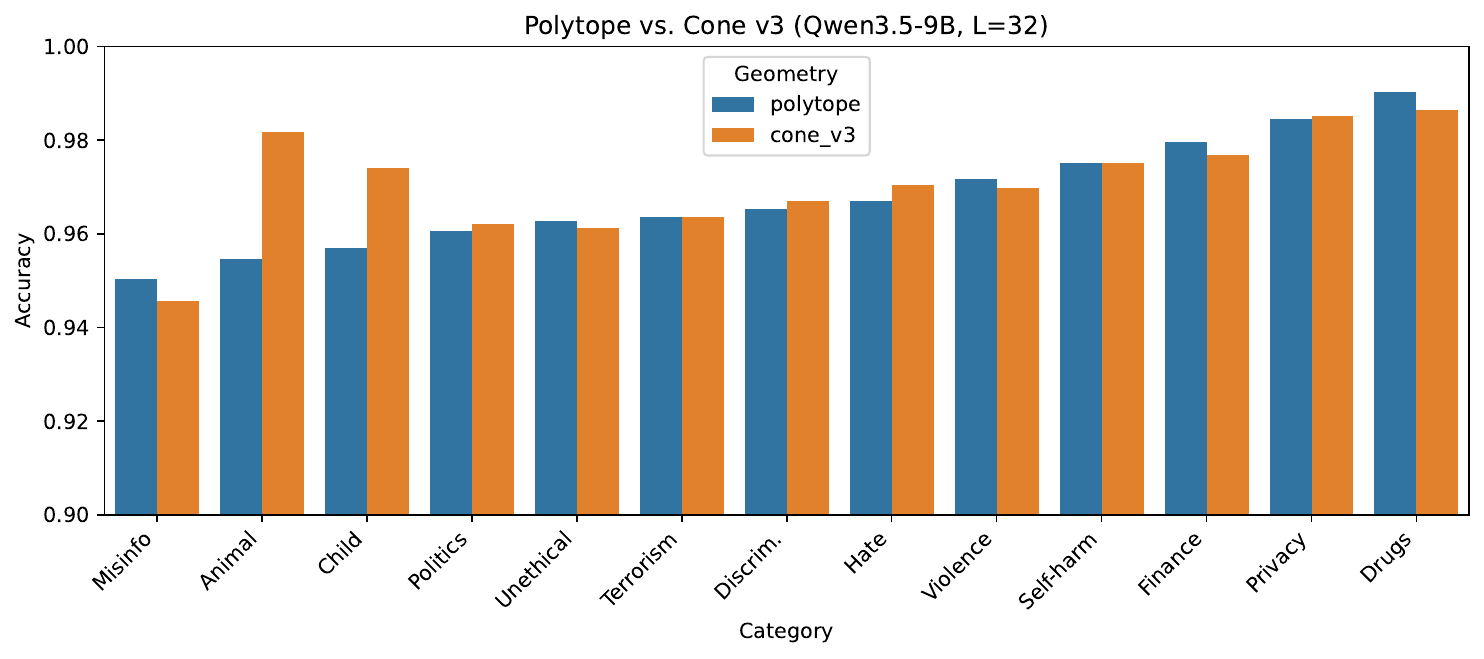}
  \caption{\poly{} vs.\ \cone{} (3-phase) accuracy (Qwen3.5-9B, L=32,
    best config per category).}
  \label{fig:geometry-comparison}
\end{figure}

\section{Per-Category Breakdown of All-Categories-Mixed Models}%
\label{app:all-mixed-percat}

This appendix supplements \S\ref{sec:all-mixed} with the per-category
test accuracy of each unified all-categories-mixed model evaluated
on the per-category balanced test splits.
The hidden-state mean and std for normalization are recomputed from
the merged training split (the same statistics used during unified
training), so each unified model is tested in the exact normalization
regime it was trained under.

\begin{table*}[t]
  \centering
  \small
  \caption{Per-category test accuracy of the unified
    all-categories-mixed models (Qwen3.5-9B, $L{=}32$, $K{=}36$).
    \cone{} is best on $9/14$ categories; \poly{} is best on $0/14$.
    \textbf{Bold} marks the per-category winner; ties resolved in
    favor of single-phase \cone{} over three-phase \cone{}.}
  \label{tab:all-mixed-percat}
  \begin{tabular}{lcccc}
    \toprule
    Category & \poly{} & \cyl{} & three-phase \cone{} & \cone{} \\
    \midrule
    animal\_abuse                          & 0.941 & \textbf{0.963} & 0.930 & 0.944 \\
    child\_abuse                           & 0.928 & 0.948 & \textbf{0.957} & 0.943 \\
    controversial\_topics,politics         & 0.866 & 0.931 & 0.935 & \textbf{0.936} \\
    discrimination,stereotype,injustice    & 0.929 & 0.940 & 0.946 & \textbf{0.954} \\
    drug\_abuse,weapons,banned\_substance  & 0.953 & 0.955 & 0.956 & \textbf{0.965} \\
    financial\_crime,property\_crime,theft & 0.947 & 0.954 & 0.955 & \textbf{0.959} \\
    hate\_speech,offensive\_language       & 0.955 & 0.946 & 0.957 & \textbf{0.964} \\
    misinformation                         & 0.907 & \textbf{0.943} & 0.941 & 0.922 \\
    non\_violent\_unethical\_behavior      & 0.943 & 0.945 & 0.949 & \textbf{0.953} \\
    privacy\_violation                     & 0.951 & 0.951 & 0.961 & \textbf{0.966} \\
    self\_harm                             & 0.950 & 0.957 & 0.961 & \textbf{0.961} \\
    sexually\_explicit,adult\_content      & 0.934 & \textbf{0.960} & 0.958 & 0.959 \\
    terrorism,organized\_crime             & 0.923 & 0.955 & \textbf{0.959} & 0.950 \\
    violence,aiding\_and\_abetting         & 0.948 & 0.951 & 0.951 & \textbf{0.960} \\
    \midrule
    \textbf{average}                       & 0.934 & 0.950 & 0.951 & \textbf{0.953} \\
    \bottomrule
  \end{tabular}
\end{table*}

The aggregate trends from Table~\ref{tab:all-mixed} hold uniformly
per category: \poly{} is the lowest accuracy on every category and
\cone{} leads the average and the modal best.
On a small minority of categories (animal\_abuse, child\_abuse,
misinformation, sexually\_explicit, terrorism), \cyl{} or three-phase
\cone{} edges out \cone{}, suggesting marginal benefit from
ensembling across geometries; on the remaining $9$ categories
\cone{} is strictly best.
The corresponding F1 ranking is identical (\cone{} 0.952,
three-phase \cone{} 0.950, \cyl{} 0.948, \poly{} 0.935 averaged),
so the conclusion is not an artifact of class imbalance.

\paragraph{Trade-offs in per-category FPR / FNR.}
\cyl{} has the lowest \emph{average} FPR ($0.019$) and highest FNR
($0.081$) per category, consistent with the lenient behaviour
attributed to its radial collapse in \S\ref{sec:all-mixed}.
\poly{} has the lowest FNR ($0.053$) but the highest FPR ($0.079$),
again consistent with its uniform under-fitting at this $K$.
\cone{} balances the two ($\text{FPR}=0.038$, $\text{FNR}=0.057$),
giving the best accuracy.

\section{Full Layer Selection Table}\label{app:layer}

\begin{table}[H]
\centering
\caption{Layer selection: mean Acc/FPR/FNR across 14 categories.
Bold: best layer per model.}
\label{tab:layer-selection}
\small
\begin{tabular}{llccc}
\toprule
Model & L & Acc & FPR & FNR \\
\midrule
Qwen2-1.5B & 4  & .812 & .189 & .188 \\
Qwen2-1.5B & 8  & .814 & .148 & .224 \\
Qwen2-1.5B & 12 & .829 & .133 & .208 \\
Qwen2-1.5B & 16 & .847 & .139 & .167 \\
Qwen2-1.5B & 20 & .828 & .133 & .211 \\
Qwen2-1.5B & 24 & .875 & .100 & .150 \\
Qwen2-1.5B & 26 & .880 & .070 & .171 \\
Qwen2-1.5B & 28 & \textbf{.890} & .087 & .133 \\
\multicolumn{5}{r}{\scriptsize $\rho{=}0.929$, $p{=}0.001$} \\
\midrule
Qwen3.5-9B & 4  & .910 & .157 & .023 \\
Qwen3.5-9B & 8  & .952 & .048 & .048 \\
Qwen3.5-9B & 12 & .931 & .086 & .052 \\
Qwen3.5-9B & 16 & .921 & .121 & .039 \\
Qwen3.5-9B & 20 & .949 & .059 & .043 \\
Qwen3.5-9B & 24 & .910 & .121 & .059 \\
Qwen3.5-9B & 28 & .964 & .050 & .023 \\
Qwen3.5-9B & 30 & \textbf{.985} & .005 & .025 \\
Qwen3.5-9B & 32 & .972 & .024 & .032 \\
\multicolumn{5}{r}{\scriptsize $\rho{=}0.661$, $p{=}0.053$} \\
\midrule
Qwen3.5-Base & 4  & \textbf{.956} & .052 & .036 \\
Qwen3.5-Base & 8  & .949 & .041 & .061 \\
Qwen3.5-Base & 12 & .928 & .091 & .052 \\
Qwen3.5-Base & 16 & .938 & .061 & .064 \\
Qwen3.5-Base & 20 & .956 & .036 & .052 \\
Qwen3.5-Base & 24 & .951 & .027 & .071 \\
Qwen3.5-Base & 28 & .951 & .046 & .052 \\
Qwen3.5-Base & 30 & .934 & .055 & .077 \\
Qwen3.5-Base & 32 & .942 & .058 & .059 \\
\multicolumn{5}{r}{\scriptsize $\rho{=}{-}0.210$, $p{=}0.587$} \\
\midrule
gpt-oss-20b & 20 & .874 & .123 & .130 \\
gpt-oss-20b & 24 & \textbf{.881} & .113 & .125 \\
\bottomrule
\end{tabular}
\end{table}

\section{Open Science}\label{app:open-science}

\paragraph{Code.}
The training pipeline, constraint geometry implementations (polytope,
cone, cylinder), SAE feature extractor, and figure generation scripts
are available in the supplementary material.
We plan to release the full codebase upon acceptance.

\paragraph{Data.}
All experiments use the publicly available BeaverTails
dataset~\cite{ji2024beavertails} via Hugging Face
(\texttt{PKU-Alignment/BeaverTails}).
Pre-extracted hidden states and trained constraint weights will be
released with the code.

\paragraph{Models.}
We use publicly available models: Qwen3.5-9B-Instruct,
Qwen3.5-9B-Base~\cite{qwen2025qwen3}, and
Qwen2-1.5B-Instruct~\cite{yang2024qwen2}.

\paragraph{Hyperparameters.}
SAE pre-training uses a 16\,384-dimensional feature space, Adam
optimizer ($\mathrm{lr}{=}10^{-3}$), batch size 256, 10 epochs, and
sparsity weight $\lambda{=}0.1$.
Cone step allocations are detailed in Table~\ref{tab:step-alloc}.

\section{Ethical Considerations}\label{app:ethics}

This work involves training classifiers to detect harmful content
categories (violence, hate speech, self-harm, etc.) using the
BeaverTails dataset, which contains examples of harmful text.
We do not generate new harmful content; all training data comes from
the existing, published dataset.
The learned constraint surfaces are designed for \emph{detection and
prevention} of harmful outputs, not for enabling them.
We note that our per-category classifiers could in principle be
repurposed to \emph{avoid} safety filters; however, the same risk
applies to any published safety classifier, and we believe the
defensive value outweighs this concern.

\section{Generative AI Disclosure}\label{app:ai}

Claude Code (Anthropic) was used for LaTeX drafting assistance,
figure generation scripting, and data analysis automation.
All experimental results, statistical analyses, and scientific
claims were produced and verified by the authors.

\end{document}